\documentclass{article}
\usepackage{spconf,amsmath,graphicx}
\usepackage{tabularx}
\usepackage{caption}
\usepackage{subcaption}
\usepackage{pifont}
\usepackage{graphicx}
\usepackage{pgfplots}
\usepackage{hyperref}
\pgfplotsset{compat=1.16}
\usepgfplotslibrary{statistics}
\usepackage{fancyhdr}

\newcommand{\cmark}{\ding{51}}%
\newcommand{\xmark}{\ding{55}}%

\newcommand\netname{6DRepNet}
\title{6D Rotation representation for unconstrained head pose estimation}
\name{Thorsten Hempel, Ahmed A. Abdelrahman, Ayoub Al-Hamadi\thanks{This research was funded by the Federal Ministry of Education and Research of Germany (BMBF) project~no.~03ZZ0448L, no.~13N16336, by the German Research Foundation (DFG) project AL 638/15, and the OVGU innovation fund.}}

\address{Faculty of Electrical Engineering and Information Technology, Neuro-Information Technology\\
Otto von Guericke University, Magdeburg, Germany}
\begin{document}
%
\maketitle
\begin{abstract}
In this paper, we present a method for unconstrained end-to-end head pose estimation. We address the problem of ambiguous rotation labels by introducing the rotation matrix formalism for our ground truth data and propose a continuous 6D rotation matrix representation for efficient and robust direct regression. This way, our method can learn the full rotation appearance which exceeds the capabilities of previous approaches that restrict the pose prediction to a narrow-angle for satisfactory results. In addition, we propose a geodesic distance-based loss to penalize our network with respect to the \textit{SO}(3) manifold geometry. Experiments on the public AFLW2000 and BIWI datasets demonstrate that our proposed method significantly outperforms other state-of-the-art methods by up to 20\%.
We open-source our training and testing code along with our trained models: \url{https://github.com/thohemp/6DRepNet}.
\end{abstract}
\begin{keywords}
head pose estimation, orientation regression, rotation matrix, geodesic loss
\end{keywords}
\section{Introduction}
\label{sec:intro}
Head pose estimation from a single image is a key task in facial analysis that is used for a wide range of applications such as driver assistance~\cite{4357803}, augmented reality~\cite{5443483}, and human-robot interaction~\cite{app11125366}. Current methods are commonly divided into landmark-based and landmark-free approaches. Landmark-based methods~\cite{8297015} detect facial landmarks in an initial step and, subsequently, recover the 3D head pose by establishing correspondence between these landmarks and a 3D head model. While this approach can lead to very accurate results, it is highly dependent on the correct prediction of the landmark positions. Hence, inferior landmark localization caused by occlusion and extreme rotation can consequently impair an accurate head pose estimation. 

Landmark-free approaches overcome this problem by directly estimating the head pose. These methods commonly facilitate deep neural networks to formulate the orientation prediction as an appearance-based task. HopeNet~\cite{Ruiz2018FineGrainedHP} presented a multi-loss approach by binning the target angle range to combine a cross-entropy and a mean squared error loss function for Euler angle prediction. Similarly, QuatNet~\cite{8444061} adapts the cross-entropy paradigm but splits classification and regression into separate network branches. One branch is used for classifying Euler angles and the second one for regressing the pose in quaternion representation. Similarly, HPE~\cite{Huang2020ImprovingHP} treats classification and regression separately and then averages the outputs as a pose regression subtask. WHENet~\cite{Zhou2020WHENetRF} keeps the single branch strategy but increases the number of bins to extend the predictable yaw range together with an EfficientNet backbone. Whereas, FSA-Net~\cite{Yang_2019_CVPR} proposes a network with a stage-wise regression and feature aggregation scheme for predicting Euler angles. TriNet~\cite{Cao_2021_WACV} adapts this method, but estimates the three unit vectors of the rotation matrix instead of Euler angles and incorporates an additional orthogonality loss to stabilize the predictions. In another approach, FDN~\cite{Zhang2020FDNFD} proposes a feature decoupling method to explicitly learn discriminative features of different head orientations.

It has become a common convention to split up the continuous rotation variables into bins for classification to stabilize the predictions~\cite{Ruiz2018FineGrainedHP,8444061,Huang2020ImprovingHP,Zhou2020WHENetRF,Zhang2020FDNFD}. However, this is problematic as pruning segments of angles into bins will consequently lead to a loss of information. Additionally, most of the current methods use the Euler angle or quaternion representation to train their networks. However, Zhou et al.~\cite{Zhou2019OnTC} demonstrated that any representation of rotation with four or fewer dimensions is discontinuous and therefore not ideal to use in a learning task for neural networks. 
\\
\\
\textbf{Contributions:} We propose a landmark-free head pose estimation method that uses the rotation matrix representation for regressing accurate head orientations. The nine-parameter matrix enables full pose regression without suffering ambiguity problems. Additionally, it allows simplifying the network by detaching unnecessary performance stabilizing measures used by other methods, \textit{e.g.} discretization of the rotation variables into a classification problem. This simplicity makes our network easily to be transferred into other rotation-related problems. Instead of predicting the entire nine-parameter rotation matrix, we efficiently regress a compressed 6D form that is transformed into the rotation matrix in a subsequent task. 

\begin{figure}
    \centering
    \includegraphics[width=\linewidth]{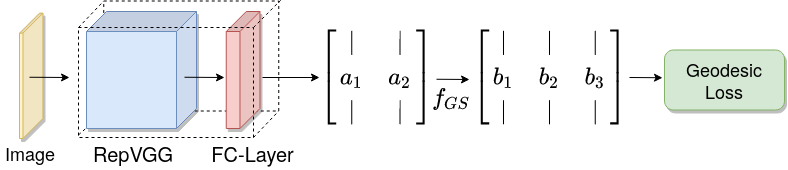}
    \caption{Overview of the proposed method.}
    \label{fig:method}
\end{figure}

Further, we propose to use the geodesic loss instead of the commonly used mean squared error loss. This way, we can use the distance angle to penalize our network in the training process that encapsulates the Special Orthogonal Group \textit{SO}(3) manifold geometry. Fig \ref{fig:method} shows an overview of your proposed method. Each component will be explained in detail in the following sections. Inspired by the 6D representation that is used in our approach, we call our network \netname.

Our training code, testing code, and trained CNN models are made publicly available to facilitate research experimentation and practical application development.

\section{Method}
A key aspect for tackling direct orientation prediction is the use of an appropriate rotation representation. A popular and convenient representation is the Euler angle. However, this representation is not optimal as it suffers from the \textit{gimbal lock} in which case there are infinitive rotation parameterizations for the same visual head pose appearance. As a consequence, neural networks have difficulties to learn the accurate pose. Whereas, on the contrary, the quaternion representation doesn't suffer from the gimbal lock, it still has an ambiguity caused by its antipodal symmetry. Especially when learning the full range of head orientations, this can lead to decreased estimation performance. A more favorable rotation representation is the rotation matrix, which is a continuous representation with a distinct parameterization for each rotation. In \textit{SO}(3) the matrix representation $R$ is sized $3\times3$ with orthogonality constraint $RR^T=I$, where $R^T$ is the transposed matrix and $I$ the identity matrix. One could now try to regress the rotation matrix directly, but this would require finding all nine parameters and enforcing the orthogonality constraint, either via the Gram-Schmidt process or by finding the nearest optimal solution using SVD. Instead, we follow the approach by Zhou et al.~\cite{Zhou2019OnTC} and perform the Gram-Schmidt mapping inside the representation itself by simply dropping the last column vector of the rotation matrix. This reduces the $3\times3$ matrix into a six parameter rotation representation that has been reported to introduce smaller errors for direct regression~\cite{Zhou2019OnTC}. 

\begin{equation}
    g_{GS}=\left( \left[ \begin{matrix}| & | & | \\ a_1 & a_2 & a_3 \\ | & | & | \\\end{matrix} \right]\right) = \left[  \begin{matrix}   | & | \\ a_1 & a_2  \\ | & |   \end{matrix}\right]
\end{equation}

The predicted 6D representation matrix can then mapped back into \textit{SO}(3).

\begin{equation}
    f_{GS}=\left( \left[\begin{matrix}| & |  \\ a_1 & a_2  \\ | & |  \\\end{matrix}  \right]\right) = \left[ \begin{matrix}| & | & | \\ b_1 & b_2 & b_3 \\ | & | & | \\\end{matrix}\right]
\end{equation}

Hereby, the remaining column vector is simply determined by the cross product that ensures that the orthogonality constraint is satisfied for the resulting $3\times3$ matrix.
\begin{equation}
\begin{split}
& b_1 = \frac{a_1}{||a_1||} \\
& b_2 = \frac{u_2}{||u_2||}, u_2=a_2-(b_1\cdot a_2)b_1\\
& b_3 = b_1\times b_2\\
\end{split}
\end{equation}

As a result, our network only has to predict 6 parameters that are mapped into a $3\times3$ rotation matrix in a subsequent transformation which at the same time also satisfies the orthogonality constraint.

A commonly used loss function for head pose related tasks is the \textit{l2}-norm. However, using the Frobenius norm for measuring distances between two matrices would break with the \textit{SO}(3) manifold geometry. Instead, the shortest path between two 3D rotations is geometrically interpreted as the geodesic distance. Let $R_p$ and $R_{gt}$ $\in$ \textit{SO}(3) be the estimated and the ground truth rotation matrices, respectively, then the geodesic distance between both rotation matrices is defined as:

\begin{equation}
    L_{g}=cos^{-1}\left( \frac{tr(R_p R_{gt}^T)-1}{2}\right)
\end{equation}

In the following, we will use this metric as the loss function for our neural network to compute accurate distance information between the predicted and the ground truth orientation.

\begin{table*}[t]
\centering
\begin{tabular} { l c@{\hskip .3in} c c c c @{\hskip .5in} c c c c }
\hline
& & \multicolumn{4}{c}{\textbf{AFLW2000}} {\hskip .5in} & \multicolumn{4}{c}{\textbf{BIWI}}\\
\hline
  & Full Range$^1$& Yaw & Pitch & Roll & MAE & Yaw & Pitch & Roll & MAE \\
  \hline
 HopeNet ($\alpha=2$) ~\cite{Ruiz2018FineGrainedHP}& \xmark  & 6.47 & 6.56 & 5.44 & 6.16  & 5.17& 6.98& 3.39 & 5.18\\ 
 HopeNet ($\alpha=1$) ~\cite{Ruiz2018FineGrainedHP}& \xmark & 6.92 & 6.64 & 5.67 & 6.41 & 4.81& 6.61& 3.27 & 4.90\\ 
 FSA-Net~\cite{Yang_2019_CVPR}& \xmark  & 4.50 & 6.08 & 4.64 & 5.07 & 4.27 & 4.96 & 2.76 & 4.00\\ 
 HPE~\cite{Huang2020ImprovingHP} & \xmark & 4.80 & 6.18 & 4.87 & 5.28 & 3.12 & 5.18 & 4.57 & 4.29 \\
 QuatNet~\cite{8444061}& \xmark  & 3.97 & 5.62  & 3.92 & 4.50 & \textbf{2.94} & 5.49 & 4.01 & 4.15  \\
 WHENet-V~\cite{Zhou2020WHENetRF}& \xmark &4.44& 5.75& 4.31& 4.83 & 3.60 & \textbf{4.10} & 2.73 & 3.48\\
 WHENet~\cite{Zhou2020WHENetRF} & \cmark \xmark & 5.11 & 6.24 & 4.92 & 5.42 & 3.99 & 4.39 & 3.06 & 3.81 \\
 TriNet~\cite{Cao_2021_WACV}& \cmark  & 4.04 & 5.77 & 4.20 & 4.67 & 4.11 & 4.76 & 3.05 & 3.97 \\
 FDN~\cite{Zhang2020FDNFD} & \xmark & 3.78 & 5.61 & 3.88 & 4.42 & 4.52 & 4.70 & \textbf{2.56} & 3.93\\
 \hline\\[-2ex] 
 \netname& \cmark &  \textbf{3.63} & \textbf{4.91} & \textbf{3.37} & \textbf{3.97} & 3.24 & 4.48 & 2.68 & \textbf{3.47}\\ 
 \hline
\end{tabular}
\caption{Comparisons with the state-of-the-art methods on the
AFLW2000 and BIWI dataset. All models are trained on the 300W-LP dataset. $^1$ These methods allow full range predictions.}
\label{table1}
\end{table*}

\section{Experiments}
\label{sec:experiments}
We implement our proposed network using Pytorch. As backbone, we choose RepVGG~\cite{Ding2021RepVGGMV}. 
For training, the RepVGG is designed as a multi-branch model just like ResNet~\cite{7780459} or Inception~\cite{7298594}. For deployment, the model can be converted into a VGG-like architecture using a re-parameterization scheme. The trimmed model yields the same accuracy but with a shorter inference time. This way, RepVGG combines the accuracy of multi-branch models with the efficiency of single branch architectures. RepVGG provides multiple architectures sizes, where we use the RepVGG-B1g2 that acts as the equivalent to the ResNet50. For the final layers, we choose a single fully connected layer with 6 outputs. As groundwork for this choice, we tested multiple configurations for the final layers, including one layer up to three sequential fully connected layers, single final layers with 6 output neurons, and separated branches with one output neuron each. In our experiments, a single final layer with 6 output neurons performed the best. The network is trained for 30 epochs using the Adam optimizer with initial learning of $1e^{-5}$ for the backbone and $1e^{-4}$ for the final fully connected layer. Both learning rates are halved every 10 epochs. We use a batch size of 64.  
\\
\\

\textbf{Datasets:} 
In general, our network is able to learn the full range of rotation. Unfortunately, due to the type of annotation technique, the most popular datasets for head pose estimation contain mainly frontal face samples.
For evaluation, we use three public available datasets: 300W-LP~\cite{Zhu2016FaceAA}, ALFW2000~\cite{7298679} and BIWI~\cite{FDGF12}. 
300W-LP consists of 66,225 face samples collected from multiple databases that are further enhanced to 122,450 samples by image flipping. It is based on around 4000 real images. The ground truth is provided in the Euler angle format. For training, we convert them into the matrix form.

The ALFW2000 dataset contains the first 2,000 images from the ALFW dataset annotated with the ground truth 3D faces and the corresponding 68 landmarks. It contains samples with large variations, different illumination, and occlusion conditions.

The BIWI dataset includes 15,678 images that were created in a lab environment with 20 participants. In this dataset, the head takes up only a small area in the images. Hence, we use the MTCNN~\cite{Zhang2016JointFD} face detector to loosely crop the heads from the images. 

For a fair comparison, we follow the preprocessing strategy from the other methods~\cite{Ruiz2018FineGrainedHP,Cao_2021_WACV} and only keep samples that have Euler angles between -99° and 99°.
\\
\\
\begin{figure}  \centering
  \begin{subfigure}{0.33\linewidth}
    \includegraphics[width=\linewidth]{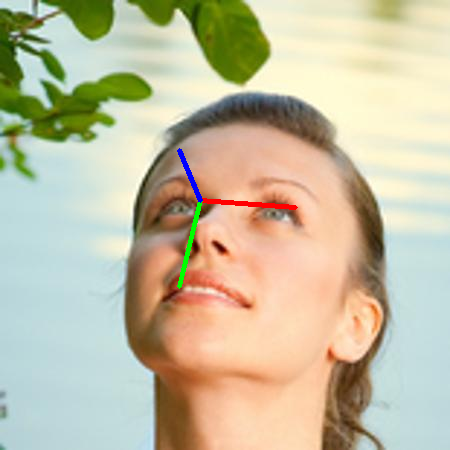}
  \end{subfigure}
  \begin{subfigure}{0.32\linewidth}
    \includegraphics[width=\linewidth]{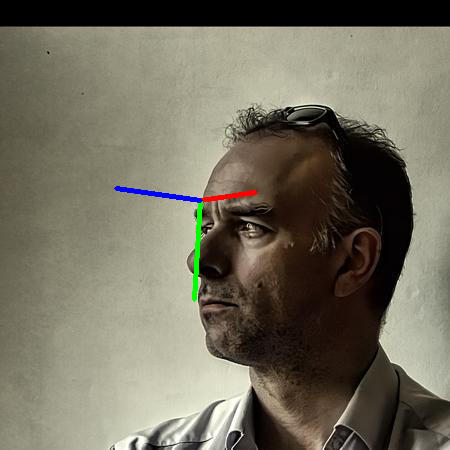}
  \end{subfigure}
  \begin{subfigure}{0.32\linewidth}
    \includegraphics[width=\linewidth]{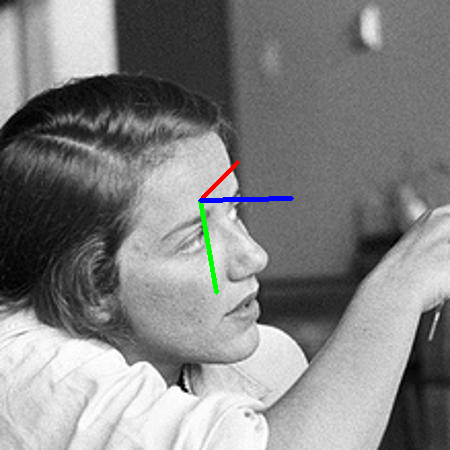}
  \end{subfigure}
  \begin{subfigure}{0.32\linewidth}
    \includegraphics[width=\linewidth]{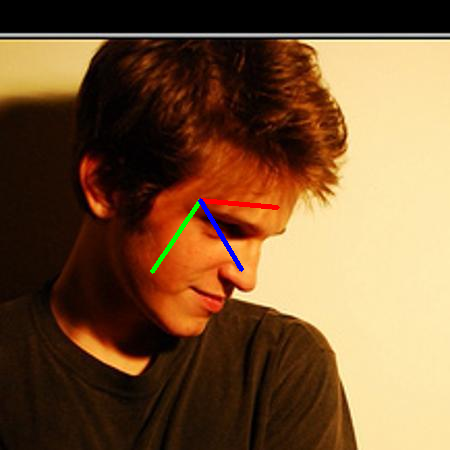}
  \end{subfigure}
    \begin{subfigure}{0.32\linewidth}
    \includegraphics[width=\linewidth]{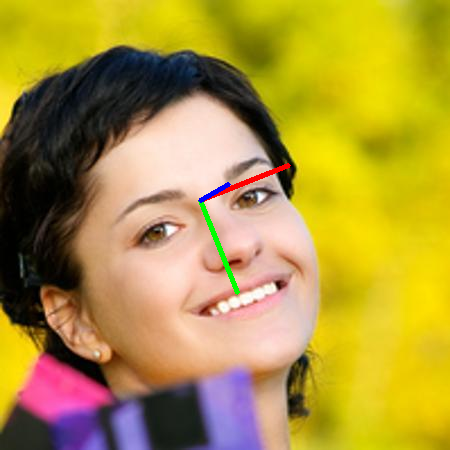}
  \end{subfigure}
    \begin{subfigure}{0.32\linewidth}
    \includegraphics[width=\linewidth]{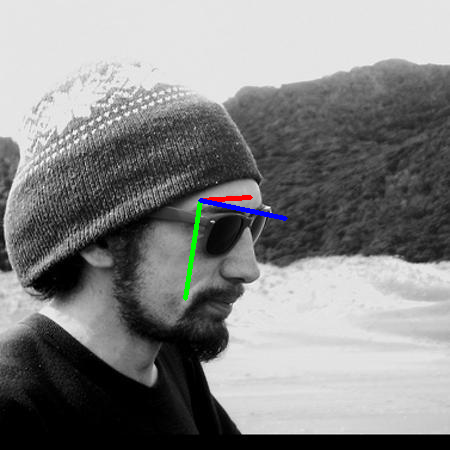}
  \end{subfigure}
  \caption{Example images with converted Euler angle visualization from the AFLW2000 dataset.}
  \label{fig:coffee}
  \label{qual}
\end{figure}
\textbf{Experiment 1:} In our first experiment, we follow the convention by using the synthetic 300W-LP dataset for training and the two real-world datasets ALFW2000 and BIWI for testing. The standard evaluation metric is the Mean Absolute Error (MAE) of the Euler angles, so we convert our rotation matrix predictions into Euler angles for better comparison. Table \ref{table1} shows the results from our first experiment setup. We compare our methods against the results reported by other state-of-the-art landmark-free approaches for head pose estimation. It demonstrates that our method outperforms the current state of the art by almost 20\% on the AFLW2000 test dataset and achieves the lowest error rate on all three kinds of rotation angles yaw, pitch, and roll. On the BIWI dataset, our method achieves state-of-the-art results for the MAE. In contrast to other methods that report very diverging results of the single angle errors, or approach reports overall very balanced errors. This indicates that our network is able to learn in a consistent and robust manner. 

For better interpretation, we added an extra column to show which methods are in general able to predict full-range rotations and which restrict their predictions within a certain range of angles. It shows that besides our approach, only two other methods target full range regression. The remaining methods have a special network architecture dedicated to narrow range head pose predictions. Most similar to our method, TriNet~\cite{Cao_2021_WACV} uses the matrix representation and directly predicts the complete matrix. As this prediction cannot be assumed to satisfy the orthogonality constraint, it needs an excessive post-process by finding an appropriate rotation matrix that is close to the prediction and at the same time has orthogonal column norm vectors. The second method marked with wide range prediction probabilities, WHENet~\cite{Zhou2020WHENetRF}, only allows the full rotation for the yaw (but not for pitch and roll) by adding more classes to their classification problem. It is noticeable that this network performs worse compared to their similar network adaptation WHENet-V that has been restricted to be only capable of predicting angles between -90° and 90°. We argue that this decline in accuracy could be caused by the introduced label ambiguities as they use the Euler representation for training. 
\\
Fig. \ref{qual} shows qualitative results from the AFLW2000 dataset. We visualized the predicted Euler angles to demonstrate how the head poses are estimated in the image samples.
\\
\\
\textbf{Experiment 2}: For our second experiment, we follow the convention by FSA-Net~\cite{Yang_2019_CVPR} and randomly split the BIWI dataset in a ratio of 7:3 for training and testing, respectively. Table \ref{table2} shows our results compared with other state-of-the-art methods that followed the same testing strategy. It demonstrates that our method outperforms all other methods not only in the overall MAE but at the same time also in predicting the yaw, pitch, and roll equally to our experiment on the AFLW2000 dataset. This supports the observed robustness in experiment 1, that achieving stable accurate results for all three angles does not only depend on the trained dataset, but rather on our proposed method itself. 
\\
\begin{table}[t]
\begin{tabularx}{\linewidth} 
{    l  @{\hskip .3in}  c    c  c   c  }
\hline
& \multicolumn{4}{c}{\textbf{BIWI}}\\
\hline
  & Yaw & Pitch & Roll & MAE \\
  \hline
 HopeNet ($\alpha =1$)~\cite{Ruiz2018FineGrainedHP} & 3.29 & 3.39 & 3.00 &  3.23\\ 
 FSA-Net~\cite{Yang_2019_CVPR}& 2.89 & 4.29 & 3.60 & 3.60 \\
 TriNet~\cite{Cao_2021_WACV} & 2.93 & 3.04 & 2.44 & 2.80 \\
 FDN~\cite{Zhang2020FDNFD} &3.00 & 3.98&  2.88&  3.29 \\
  \hline\\[-2ex] 
\netname & \textbf{2.69} & \textbf{2.92} & \textbf{2.36} & \textbf{2.66}\\
\hline
\end{tabularx}
\caption{Comparisons with the state-of-the-art methods on the
BIWI dataset. 70\% of the BIWI dataset is used for training and the remaining 30\% for testing.}
\label{table2}
\end{table}
\begin{table}[t]
\begin{tabularx}{\linewidth} 
{    l  @{\hskip .5in}  c   c  c  }
\hline

& \textbf{AFLW2000} &\textbf{BIWI} & \textbf{70/30 BIWI}\\
\hline
  & MAE  & MAE & MAE \\
  \hline
$\ell_2$ &  4.13 & 3.70 & 2.84 \\
Geodesic & 3.97 & 3.47 & 2.66\\

\hline
\end{tabularx}
\caption{Comparison of the MAE between $\ell_2$ and geodesic loss.}
\label{table3}
\end{table}
\\
\textbf{Experiment 3:} Most current methods use the $\ell_2$-norm for calculating the loss in the training procedure. We argue that the Geodesic distance is a better distance metric to measure the prediction accuracy for head pose orientation. To prove this, we conduct another experiment where we repeat our previous tests, but this time train our network with the  $\ell_2$ distance loss. Table \ref{table3} shows these results compared to our models trained with geodesic distance loss. It states that the networks with geodesic loss penalty performed slightly better than those that were trained with  $\ell_2$-norm. 
\\
\\
\textbf{Experiment 4:} In a final experiment, we analyze the impact of the chosen backbone on the results. ResNet50 is a popular standard network that is also used by HopeNet and TriNet, so we use it as the comparison backbone. Table \ref{table4} shows that our method with RepVGG backbone is capable to perform about 7\% better on all test scenarios against the $\ell_2$ loss. However, with the aid of the ResNet50 backbone, our method would still achieve state-of-the-art results on the AFLW2000 dataset.

\begin{table}[]
\begin{tabularx}{\linewidth} 
{    l  @{\hskip .2in}  c   c  c  }
\hline
& \textbf{AFLW2000} &\textbf{BIWI} & \textbf{70/30 BIWI}\\
\hline
& MAE  & MAE & MAE \\
  \hline
ResNet50 & 4.26 & 3.76 & 3.09\\
RepVGG-B1g2 &  3.97 & 3.47 & 2.66 \\

\hline
\end{tabularx}
\caption{Comparison of the MAE between the different backbones.}
\label{table4}
\end{table}

\section{Conclusion}
In this paper, we presented a method for unconstrained end-to-end head pose estimation from single images. We follow the argument that the rotation matrix is more suitable for orientation learning tasks and propose a continuous 6D rotation matrix representation for efficient direct regression. Additionally, we introduce the geodesic loss instead of the commonly used MSE for the robust training.
Also different from previous approaches, our method can regress full rotations and does not utilize an angle restricting binning principle. Nonetheless, our method outperforms other state-of-the-art methods on multiple datasets by up to 20\%. In additional experiments, we analyze the impact of the backbone and loss function on our results.
In the future, we aim to draw on our method's full potential by training it on a dataset that provides full rotation samples.
To this end, a potential dataset has been introduced by WHENet~\cite{Zhou2020WHENetRF} that presented a method for recovering full head pose annotations for the CMU Panoptic dataset~\cite{jooiccv2015}.





\bibliographystyle{IEEEbib}
\bibliography{bib}

\begin{thebibliography}{10}

\bibitem{4357803}
Erik Murphy-Chutorian, Anup Doshi, and Mohan~Manubhai Trivedi,
\newblock ``Head pose estimation for driver assistance systems: A robust
  algorithm and experimental evaluation,''
\newblock in {\em 2007 IEEE Intelligent Transportation Systems Conference},
  2007, pp. 709--714.

\bibitem{5443483}
Erik Murphy-Chutorian and Mohan~Manubhai Trivedi,
\newblock ``Head pose estimation and augmented reality tracking: An integrated
  system and evaluation for monitoring driver awareness,''
\newblock {\em IEEE Transactions on Intelligent Transportation Systems}, vol.
  11, no. 2, pp. 300--311, 2010.

\bibitem{app11125366}
Dominykas Strazdas, Jan Hintz, and Ayoub Al-Hamadi,
\newblock ``Robo-hud: Interaction concept for contactless operation of
  industrial cobotic systems,''
\newblock {\em Applied Sciences}, vol. 11, no. 12, 2021.

\bibitem{8297015}
Philipp Werner, Frerk Saxen, and Ayoub Al-Hamadi,
\newblock ``Landmark based head pose estimation benchmark and method,''
\newblock in {\em 2017 IEEE International Conference on Image Processing
  (ICIP)}, 2017, pp. 3909--3913.

\bibitem{Ruiz2018FineGrainedHP}
Nataniel Ruiz, Eunji Chong, and James~M. Rehg,
\newblock ``Fine-grained head pose estimation without keypoints,''
\newblock {\em 2018 IEEE/CVF Conference on Computer Vision and Pattern
  Recognition Workshops (CVPRW)}, pp. 2155--215509, 2018.

\bibitem{8444061}
Heng-Wei Hsu, Tung-Yu Wu, Sheng Wan, Wing~Hung Wong, and Chen-Yi Lee,
\newblock ``Quatnet: Quaternion-based head pose estimation with multiregression
  loss,''
\newblock {\em IEEE Transactions on Multimedia}, vol. 21, no. 4, pp.
  1035--1046, 2019.

\bibitem{Huang2020ImprovingHP}
Bin Huang, Renwen Chen, Wang Xu, and Qinbang Zhou,
\newblock ``Improving head pose estimation using two-stage ensembles with top-k
  regression,''
\newblock {\em Image Vis. Comput.}, vol. 93, pp. 103827, 2020.

\bibitem{Zhou2020WHENetRF}
Yijun Zhou and James Gregson,
\newblock ``Whenet: Real-time fine-grained estimation for wide range head
  pose,''
\newblock in {\em 31st British Machine Vision Conference 2020, {BMVC} 2020,
  Virtual Event, UK, September 7-10, 2020}. 2020, {BMVA} Press.

\bibitem{Yang_2019_CVPR}
Tsun-Yi Yang, Yi-Ting Chen, Yen-Yu Lin, and Yung-Yu Chuang,
\newblock ``Fsa-net: Learning fine-grained structure aggregation for head pose
  estimation from a single image,''
\newblock in {\em Proceedings of the IEEE/CVF Conference on Computer Vision and
  Pattern Recognition (CVPR)}, June 2019.

\bibitem{Cao_2021_WACV}
Zhiwen Cao, Zongcheng Chu, Dongfang Liu, and Yingjie Chen,
\newblock ``A vector-based representation to enhance head pose estimation,''
\newblock in {\em Proceedings of the IEEE/CVF Winter Conference on Applications
  of Computer Vision (WACV)}, January 2021, pp. 1188--1197.

\bibitem{Zhang2020FDNFD}
Hao Zhang, Mengmeng Wang, Yong Liu, and Yi~Yuan,
\newblock ``Fdn: Feature decoupling network for head pose estimation,''
\newblock in {\em AAAI}, 2020.

\bibitem{Zhou2019OnTC}
Yi~Zhou, Connelly Barnes, Jingwan Lu, Jimei Yang, and Hao Li,
\newblock ``On the continuity of rotation representations in neural networks,''
\newblock {\em 2019 IEEE/CVF Conference on Computer Vision and Pattern
  Recognition (CVPR)}, pp. 5738--5746, 2019.

\bibitem{Ding2021RepVGGMV}
Xiaohan Ding, X.~Zhang, Ningning Ma, Jungong Han, Guiguang Ding, and Jian Sun,
\newblock ``Repvgg: Making vgg-style convnets great again,''
\newblock {\em 2021 IEEE/CVF Conference on Computer Vision and Pattern
  Recognition (CVPR)}, pp. 13728--13737, 2021.

\bibitem{7780459}
Kaiming He, Xiangyu Zhang, Shaoqing Ren, and Jian Sun,
\newblock ``Deep residual learning for image recognition,''
\newblock in {\em 2016 IEEE Conference on Computer Vision and Pattern
  Recognition (CVPR)}, 2016, pp. 770--778.

\bibitem{7298594}
Christian Szegedy, Wei Liu, Yangqing Jia, Pierre Sermanet, Scott Reed, Dragomir
  Anguelov, Dumitru Erhan, Vincent Vanhoucke, and Andrew Rabinovich,
\newblock ``Going deeper with convolutions,''
\newblock in {\em 2015 IEEE Conference on Computer Vision and Pattern
  Recognition (CVPR)}, 2015, pp. 1--9.

\bibitem{Zhu2016FaceAA}
Xiangyu Zhu, Zhen Lei, Xiaoming Liu, Hailin Shi, and S.~Li,
\newblock ``Face alignment across large poses: A 3d solution,''
\newblock {\em 2016 IEEE Conference on Computer Vision and Pattern Recognition
  (CVPR)}, pp. 146--155, 2016.

\bibitem{7298679}
Xiangyu Zhu, Zhen Lei, Junjie Yan, Dong Yi, and Stan~Z. Li,
\newblock ``High-fidelity pose and expression normalization for face
  recognition in the wild,''
\newblock in {\em 2015 IEEE Conference on Computer Vision and Pattern
  Recognition (CVPR)}, 2015, pp. 787--796.

\bibitem{FDGF12}
G.~Fanelli, M.~Dantone, J.~Gall, A.~Fossati, and L.~van Gool,
\newblock ``Random forests for real time 3d face analysis,''
\newblock {\em International Journal of Computer Vision}, vol. 101, no. 3, pp.
  437--458, 2013.

\bibitem{Zhang2016JointFD}
Kaipeng Zhang, Zhanpeng Zhang, Zhifeng Li, and Yu~Qiao,
\newblock ``Joint face detection and alignment using multitask cascaded
  convolutional networks,''
\newblock {\em IEEE Signal Processing Letters}, vol. 23, pp. 1499--1503, 2016.

\bibitem{jooiccv2015}
Hanbyul Joo, Hao Liu, Lei Tan, Lin Gui, Bart Nabbe, Iain Matthews, Takeo
  Kanade, Shohei Nobuhara, , and Yaser Sheikh,
\newblock ``Panoptic studio: A massively multiview system for social motion
  capture,''
\newblock 2015.

\end{thebibliography}

\end{document}